%% file: main.tex
\documentclass{article}

\newif\ifneuripssubmission
\neuripssubmissionfalse %

\PassOptionsToPackage{numbers,sort}{natbib} %

\ifneuripssubmission
  \usepackage[main]{neurips_2026}
\else
  \usepackage[preprint]{neurips_2026}
\fi

\usepackage[utf8]{inputenc} %
\usepackage[T1]{fontenc}    %
\usepackage{hyperref}       %
\usepackage{url}            %
\usepackage{booktabs}       %
\usepackage{amsfonts}       %
\usepackage{nicefrac}       %
\usepackage{microtype}      %
\usepackage{xcolor}         %

\usepackage{graphicx}
\usepackage{subcaption}
\usepackage{algorithm}
\usepackage[noEnd,commentColor=black]{algpseudocodex}
\usepackage[most]{tcolorbox}
\usepackage{float}
\usepackage{wrapfig}
\usepackage{lipsum}
\usepackage{colortbl}
\usepackage{soul}
\usepackage{enumitem}
\usepackage{multirow}
\usepackage{makecell}
\usepackage{tablefootnote}
\usepackage[para,online,flushleft]{threeparttable}
\usepackage[nameinlink,noabbrev]{cleveref}
\usepackage{siunitx}

\definecolor{myblue}{rgb}{1,0.5,0}
\definecolor{urlteal}{HTML}{0F766E}
\hypersetup{
    colorlinks = true,
    linkcolor = {magenta},
    citecolor = {magenta},
    urlcolor=magenta,
}

\setlist{label=\textbullet}
\newcommand{\oursname}{FASTER}
\newcommand{\ours}{\textcolor{orange}{{\texttt{\oursname}}}}
\newcommand{\oursplain}{{\texttt{\oursname}}}
\newcommand{\oursexpo}{\textcolor{orange}{{\texttt{\oursname-EXPO}}}}
\newcommand{\oursidql}{\textcolor{orange}{{\texttt{\oursname-IDQL}}}}
\newcommand{\nonsubmissiononly}[1]{\ifneuripssubmission\else#1\fi}

\title{{\huge \texttt{\oursname:}}Value-Guided Sampling for Fast RL}

\author{
  Perry Dong\thanks{Equal contribution.}\:\: \quad Alexander Swerdlow\footnotemark[1]\:\: \quad Dorsa Sadigh \quad Chelsea Finn \\
  Stanford University \\
  \texttt{\{perryd, aswerdlo\}@stanford.edu}
}

\ifneuripssubmission
  \setlist[itemize]{topsep=2pt,itemsep=1pt,parsep=0pt,partopsep=0pt,leftmargin=*}
  \setlist[enumerate]{topsep=2pt,itemsep=1pt,parsep=0pt,partopsep=0pt,leftmargin=*}
  \makeatletter
  \renewcommand{\@toptitlebar}{
    \hrule height 4\p@
    \vskip 0.16in
    \vskip -\parskip%
  }
  \renewcommand{\@bottomtitlebar}{
    \vskip 0.18in
    \vskip -\parskip
    \hrule height 1\p@
    \vskip 0.05in%
  }
  \makeatother
\fi

\begin{document}

\maketitle

\begin{abstract}
Some of the most performant reinforcement learning algorithms today can be prohibitively expensive as they use test-time scaling methods such as sampling multiple action candidates and selecting the best one. In this work, we propose \ours{}, a method for getting the benefits of sampling-based test-time scaling of diffusion-based policies without the computational cost by tracing the performance gain of action samples back to earlier in the denoising process. Our key insight is that we can model the denoising of multiple action candidates and selecting the best one as a Markov Decision Process (MDP) where the goal is to progressively filter action candidates before denoising is complete. With this MDP, we can learn a policy and value function in the denoising space that predicts the downstream value of action candidates in the denoising process and filters them while maximizing returns. The result is a method that is lightweight and can be plugged into existing generative RL algorithms. Across challenging long-horizon manipulation tasks in online and batch-online RL, \ours{} consistently improves the underlying policies and achieves the best overall performance among the compared methods. Applied to a pretrained VLA, \ours{} achieves the same performance while substantially reducing training and inference compute requirements. %

\vspace{3pt}
Code: \href{https://github.com/alexanderswerdlow/faster}{\texttt{https://github.com/alexanderswerdlow/faster}}

\end{abstract}

\input{introduction}
\input{related_work}
\input{preliminaries}
\input{method}
\input{experiments}

\input{discussion}

\nocite{dong2025reinforcement,dong2026tql}

\newpage
\bibliographystyle{unsrtnat} %
\bibliography{references}

\clearpage
\appendix
\input{appendix.tex}

\clearpage

\end{document}

%% file: introduction.tex
\section{Introduction} \label{sec:introduction}
Recent reinforcement learning (RL) methods have demonstrated strong performance using expressive policy backbones such as diffusion models which are widely used in domains such as image/video generation and robotics. However, some of the most performant of these algorithms are prohibitively expensive both during training and at test time. This is because many of these algorithms rely on sampling multiple candidate actions and selecting the best one, for example as determined by which one has the highest value. While effective, such approaches incur substantial computational costs. This is especially problematic for large models like modern Vision-Language-Action (VLA) models, which already operate near the limits of acceptable latency. A modest sample count of eight will multiply inference costs accordingly and can render these methods impractical in latency-sensitive or resource-constrained settings. In this work, we ask: can we recover the benefits of sampling-based test-time scaling without suffering its computational cost?

\begin{figure}[t]
  \centering
  \includegraphics[width=\linewidth]{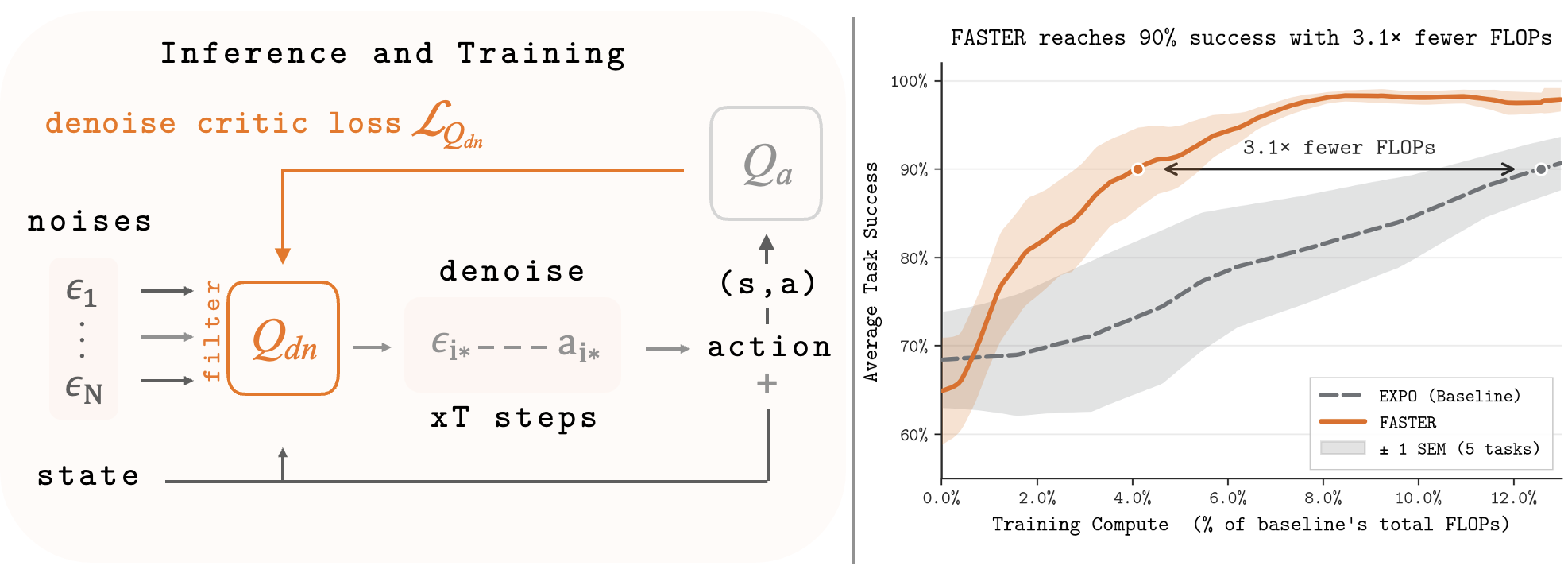}
  \caption{\footnotesize \textbf{Left: Overview of \ours{}.} Instead of denoising all $N$ candidates and selecting the best action post-hoc (best-of-$N$), \ours{} learns a denoise critic $Q^{dn}$ that scores action samples \emph{during} denoising, often directly on the initial noise itself. \textbf{Right: Performance of \ours{} compared with baseline} averaged across tasks for $\pi_{0.5}$. The FLOPs are normalized on the x-axis against the amount of FLOPs needed for the baseline to converge.}
  \label{fig:method}
\end{figure}
This high computational cost comes from policies with generative models such as diffusion or flow matching that need to denoise all candidates. Prior works have shown that test-time scaling methods for generative models are performant. Methods such as best-of-$N$ sampling and self-consistency sample multiple candidates and select the one that maximizes a given metric. While these methods are effective at maximizing performance, they scale poorly and require full denoising over all action candidates. Distillation-based approaches amortize this cost by training the policy to directly reproduce high-value behaviors, but require training a separate policy which can be expensive. The core difficulty underlying all of these approaches is that, without observing the fully denoised samples, it is hard to know which ones are worth executing since there is no direct supervision linking a given noise sample to its downstream outcome, and the mapping from noise to action is only defined implicitly through the denoising process.

In this work, our key insight is that we can model the denoising of multiple candidate samples and the selection of the best one as a Markov Decision Process (MDP) where the goal is to progressively filter candidates before denoising is complete. With this MDP, we can learn a denoise Q-function and policy with traditional temporal difference learning that decide which actions to keep and remove while maximizing the returns. To supervise the value function without requiring exhaustive rollouts, we use the value function over final actions as the reward of the action filtering MDP, enabling the model to propagate outcome information back to the denoising space without direct outcome labels for every candidate. Practically, we find that filtering candidates at the noise level—which is the cheapest instantiation computationally—achieves performance equivalent to fully denoising all action samples and selecting the highest value action. The computational cost of denoising—which is the dominant bottleneck at inference time—is thus reduced to that of a single rollout, regardless of the number of action candidates. The result is a lightweight, modular approach that decouples the benefits of broad action sample selection from the cost of full policy execution.

Our main contribution is a framework for obtaining the performance gains of sampling and selecting the best of multiple actions of RL methods without the computational overhead of denoising all actions. Unlike approaches that operate on actions, our method operates directly on the denoising process to filter down candidates, making selection both cheap and modular. We evaluate our method on challenging manipulation benchmarks, including Robomimic and LIBERO, across algorithms and model capacities, demonstrating consistent gains over single-sample baselines and near-parity with best-of-$N$ methods at a fraction of the computational overhead.

%% file: related_work.tex
\section{Related Work}
\label{sec:related_work}

\paragraph{Test-time scaling and best-of-$N$ inference.}
Significant work has shown test-time scaling methods for generative models to be highly performant. These methods take several forms, including iterative refinement, but the most common involve sampling multiple candidates and performing filtering or refinement based on a group of samples. Best-of-$N$ sampling is the most standard of such methods, with extensive use in many domains including in autoregressive language models \citep{wang2022selfconsistency,snell2024scalingllm,chow2024inferenceaware}, diffusion models \citep{ma2025itsdiffusion}, and policy learning \citep{dong2025expo, hansen2023idql, dong2026valueflows, chen2023offline}.

\paragraph{Diffusion policies and value-guided sampling.}
Diffusion models have emerged as a strong policy class for a wide variety of domains because they can represent rich, multimodal data distributions~\citep{dong2025expo, wang2022diffusionql,chi2023diffusionpolicy,hansen2023idql, yang2023policyrepresentationdiffusionprobability, park2025flow,wagenmaker2025posterior,psenka2024qsm}. Several recent works study how to improve these policies with value functions or reduce their inference cost. Value-Guided Policy Steering (V-GPS) re-ranks actions from frozen robot policies with a learned value function~\citep{nakamoto2024steeringgeneralists}, RoboMonkey scales test-time sampling and verification for vision-language-action policies~\citep{kwok2025robomonkey}, and EXPO couples an expressive base policy with a lightweight edit policy and selects value-maximizing actions online~\citep{dong2025expo}. Other work amortizes or changes the policy itself: One-Step Diffusion Policy distills multi-step denoising into a faster actor~\citep{wang2024onestep}, while DSRL post-trains a behavior-cloned diffusion policy by running RL in its latent-noise space~\citep{wagenmaker2025latentrl}. Compared to these approaches, our method learns a separate critic over \emph{which actions to filter during denoising}, enabling candidate selection before actions are denoised. As such, our method can be directly applied to methods that can benefit from sampling multiple actions. 

\paragraph{Initial-noise and noise-space methods for diffusion models.}
A rapidly growing body of literature shows that the initial seed of a diffusion model has a significant effect on generation quality. Prior work has shown that seed quality is linked to inversion stability~\citep{ban2024crystalball,qi2024notallnoises} and several methods explicitly optimize or learn better starting noises, including InitNO and FIND~\citep{guo2024initno,chen2024find}, as well as plug-and-play noise refinement methods such as Golden Noise~\citep{zhou2024goldennoise}, NoiseRefine~\citep{ahn2024noiseworth}, and Noise-Level Guidance~\citep{mannering2025nlg}. Noise Hypernetworks likewise seek to amortize test-time noise optimization into a learned auxiliary module~\citep{eyring2025noisehyper}. More closely related, TTSnap~\citep{yu2025ttsnap} prunes candidate trajectories early by decoding intermediate estimates and scoring them with noise-aware reward models trained via self-distillation, allowing it to search over larger candidate sets. This is similar in spirit to our goal of recovering the benefits of best-of-$N$ without denoising every candidate, but the mechanism is distinct. TTSnap still performs partial denoising and reward evaluation at selected timesteps, and is studied in text-to-image generation with image-reward objectives from pretrained models. By contrast, we frame the problem of filtering action candidates as an MDP and learn a \emph{state-conditioned} critic that predicts downstream value directly from the initial noise of a diffusion policy in the context of reinforcement learning.

%% file: preliminaries.tex
\section{Preliminaries} \label{sec:prelim}

We consider problems modeled as Markov decision processes (MDPs), each defined by a tuple $\{\mathcal{S}, \mathcal{A}, r, \gamma, T, \rho\}$ where $\mathcal{S}$ is the state space, $\mathcal{A}$ is the action space, $r: \mathcal{S} \times\mathcal{A} \rightarrow \mathbb{R}$ is the reward function, $T(s'|s, a)$ comprises the transition dynamics, $\gamma \in [0, 1]$ is the discount factor, and $\rho(s)$ is the initial state distribution. In this MDP, the RL agent observes state $s_{t}$ at every timestep $t$ and chooses action $a_t$ from its policy $\pi(a_{t}|s_{t})$. The tuple of states, actions, rewards, and next states $(s, a, r, s')$ is appended to a replay buffer $D$ for training. The goal of RL is to maximize the expected sum of discounted returns $\mathbb{E}_{\pi}[\sum_{t=0}^{T}\gamma^t r(s_t, a_t)]$.

In this paper, we study the setting where the policy $\pi$ is a diffusion or flow policy that requires denoising. We further focus on algorithms that perform test-time scaling through sampling multiple action candidates. Given $N$ initial noise seeds $\{\epsilon_i\}_{i=1}^N$, the policy denoises them into $N$ actions $\{a_i\}_{i=1}^N$. The policy then selects the best action among the candidate set, for example by choosing $i^* = \arg\max_i Q(s,a_i)$ and executing $a_{i^*}$. Our main goal is to capture the performance gain from these sampling-based test-time scaling approaches without the computational cost of denoising all $N$ action candidates.

%% file: method.tex
\section{\oursplain} \label{sec:method}
To capture the benefits of test-time scaling without its computational overhead, we leverage a key insight: the sample variance that drives these gains can be traced to early in the denoising process. We introduce \ours{}, a framework that models denoising as an MDP and learns a filtering policy to select promising actions before they are denoised. An overview is shown in \Cref{fig:method}. We first define the filtering MDP (\Cref{subsec:filter}), then describe policy learning within this MDP (\Cref{subsec:policy}). In practice, we find that sample variance is largely determined by the initial noise, motivating a practical instantiation that filters at the noise level — the earliest and computationally cheapest point in the denoising chain. We describe this and further practical implementations in \Cref{sec:practical_implementation}.

\subsection{Modeling Action Denoising as an MDP} \label{subsec:filter}

\begin{wrapfigure}[17]{R}{0.38\textwidth}
    \vspace{-1cm}
    \centering
    \includegraphics[width=0.4\columnwidth]{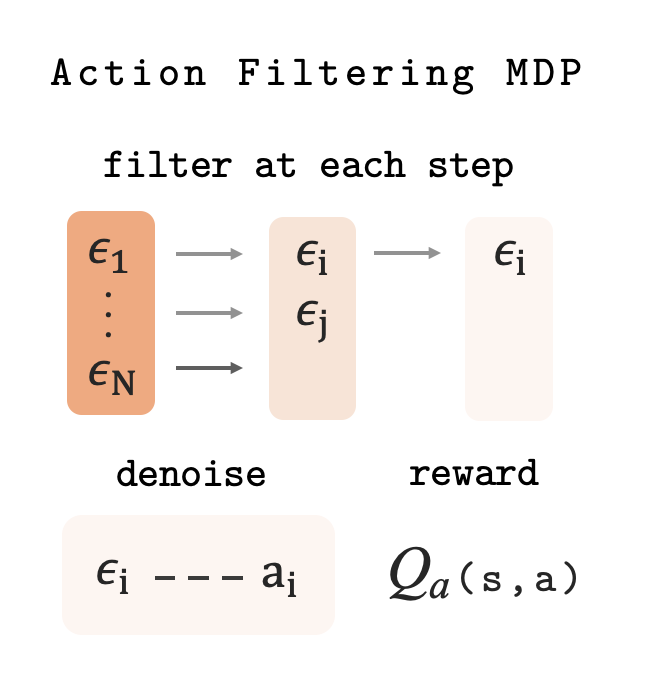}
    \caption{
        \footnotesize 
        \textbf{Action Filtering MDP. } We model the process of denoising action candidates and selecting the best one as an MDP where the goal is to filter action samples during denoising while maximizing returns.}
    \label{fig:mdp}
\end{wrapfigure}

We want to select the best action candidates \emph{before} denoising completes. To do so, we model the reverse diffusion process as a filtering MDP over $N$ noise candidates $\{\epsilon_i\}_{i=1}^N$, $\epsilon_i \sim \mathcal{N}(0,I)$, where a policy progressively discards unpromising candidates at each denoising step. We illustrate this in Figure \ref{fig:mdp}. 

\textbf{States.} A state $\mathbf{s}_t = (s, t, \mathcal{C}_t, \{a_i^{(t)}\}_{i \in \mathcal{C}_t})$ consists of the environment state $s$, the denoising timestep $t \in \{T, \dots, 1\}$, the surviving candidate set $\mathcal{C}_t \subseteq \{1, \dots, N\}$, and the partially denoised intermediates, from pure noise $a_i^{(T)} = \epsilon_i$ to a clean action at $t{=}1$.

\textbf{Actions.} At each step the policy retains or discards each candidate: $m_{t,i} \in \{0, 1\}$ for $i \in \mathcal{C}_t$, with at least one retained ($\sum_i m_{t,i} \geq 1$). The surviving set becomes $\mathcal{C}_{t-1} = \{i : m_{t,i} = 1\}$.

\textbf{Transitions.} A denoising step is applied to each candidate, producing $a_i^{(t-1)}$. The episode ends when a single candidate remains ($|\mathcal{C}_{t-1}| = 1$) or denoising finishes ($t = 1$), at which point the highest-scoring survivor is executed. The episode may terminate at $t > 1$, in which case the remaining candidate is denoised until $t=1$.

\textbf{Reward.} Non-terminal steps receive zero reward. At termination the reward is $Q^a(s, a_{i^*})$, where $a_{i^*}$ is the executed action.\looseness=-1

\subsection{Learning the filtering policy} \label{subsec:policy}

To learn the filtering policy in the MDP above, we learn an action-value function over filtering decisions. Concretely, we learn the denoise critic $Q^{dn}(\mathbf{s}_t, \mathbf{m}_t)$, where $\mathbf{s}_t = (s, t, \mathcal{C}_t, \{a_i^{(t)}\}_{i \in \mathcal{C}_t})$ is the filtering state and $\mathbf{m}_t = \{m_{t,i}\}_{i \in \mathcal{C}_t}$ is the filtering action with $m_{t,i} \in \{0,1\}$ denoting whether candidate $i$ is retained.

We train $Q^{dn}$ with a temporal-difference (TD) objective.

\begin{equation}
\mathcal{L}_{\mathrm{TD}} = \mathbb{E}_{(\mathbf{s}_t,\mathbf{m}_t,\mathbf{s}_{t-1})}\left[\left(Q^{dn}(\mathbf{s}_t, \mathbf{m}_t) - \left[r_t + \gamma \max_{\mathbf{m}_{t-1}} \bar{Q}^{dn}(\mathbf{s}_{t-1}, \mathbf{m}_{t-1})\right]\right)^2\right]
\end{equation}

where $r_t = 0$ for non-terminal transitions and, at termination, $r_t = Q^a(s, a_{i^*})$ for the executed action $a_{i^*}$ induced by the final surviving candidate.

At test time, we sample $N$ noise candidates $\{\epsilon_i\}_{i=1}^N$ with $\epsilon_i \sim \mathcal{N}(0, I)$ and repeatedly apply the filtering policy induced by $Q^{dn}$ to choose filtering actions $\mathbf{m}_t$ until a single candidate remains. Let $i^*$ denote the surviving candidate index. That survivor is then denoised to obtain the final action $a_{i^*} = \pi_\theta(s, \epsilon_{i^*})$. Depending on the method (\Cref{sec:practical_implementation}), we either use $a_{i^*}$ as the final action or consider this alongside a proposal action from an edit policy.

\subsection{Practical implementation} \label{sec:practical_implementation}

\begin{figure*}[t]
  \centering
  \begin{minipage}[t]{0.485\textwidth}
    \begin{algorithm}[H]
      \caption{Learning procedure in \ours{}}\label{alg:ours_learning}
      \begin{algorithmic}[1]
        \Require $s$, $\pi_\theta$, $Q^{dn}$, $Q^a$, $N$
        \State Sample $\{\epsilon_i\}_{i=1}^N$ with $\epsilon_i \sim \mathcal{N}(0, I)$
        \State $i^* \leftarrow \arg\max_i Q^{dn}(s, \epsilon_i)$
        \State $a_{i^*} \leftarrow \pi_\theta(s, \epsilon_{i^*})$
        \State $\mathcal{L}_{\mathrm{reg}} = \left\| Q^{dn}(s, \epsilon_{i^*}) - \mathrm{sg}[Q^a(s, a_{i^*})] \right\|_2^2$
      \end{algorithmic}
    \end{algorithm}
  \end{minipage}\hfill
  \begin{minipage}[t]{0.485\textwidth}
    \begin{algorithm}[H]
      \caption{Inference procedure in \ours{}}\label{alg:ours_inference}
      \begin{algorithmic}[1]
        \Require $s$, $\pi_\theta$, $Q^{dn}$, $N$
        \State Sample $\{\epsilon_i\}_{i=1}^N$ with $\epsilon_i \sim \mathcal{N}(0, I)$
        \State $i^* \leftarrow \arg\max_i Q^{dn}(s, \epsilon_i)$
        \State $a_{i^*} \leftarrow \pi_\theta(s, \epsilon_{i^*})$
        \State Execute $a_{i^*}$
      \end{algorithmic}
    \end{algorithm}
  \end{minipage}
\end{figure*}

In our setting, we restrict the MDP to only filter at $t=T$, reducing the multi-step filtering process to a single decision. Since the filtering happens before any denoising, the state collapses to just the environment state $s$ (the denoising timestep is fixed at $T$ and there is no history of prior filtering steps), and the filtering ``action'' reduces to selecting a single candidate rather than producing a binary mask $\mathbf{m}_t$ over survivors. Under this simplification, the joint $Q^{dn}(\mathbf{s}_t, \mathbf{m}_t)$ from \Cref{subsec:policy} decomposes into per-candidate scores: we learn a \emph{noise-level critic} $Q^{dn}(s, \epsilon)$ that maps an environment state $s$ and a single noise sample $\epsilon \sim \mathcal{N}(0, I)$ to the expected return of denoising that sample. We denote the standard action-space critic as the \emph{action-level critic} $Q^{a}(s, a)$, which provides the supervision signal. We use a greedy filtering policy which selects $i^* = \arg\max_i Q^{dn}(s, \epsilon_i)$ and denoises $\epsilon_{i^*}$ into $a_{i^*} = \pi_\theta(s, \epsilon_{i^*})$.

Because the MDP is now one step, the TD objective in \Cref{subsec:policy} simplifies to a one-step regression against the action-level critic. We sample $N$ noise candidates $\{\epsilon_i\}_{i=1}^N$ with $\epsilon_i \sim \mathcal{N}(0, I)$, select the highest-scoring index $i^* = \arg\max_{i} Q^{dn}(s, \epsilon_i)$, denoise $\epsilon_{i^*}$ to obtain $a_{i^*} = \pi_\theta(s, \epsilon_{i^*})$, and supervise $Q^{dn}$ by regressing to the action-level critic:
\[
\mathcal{L}_{\mathrm{reg}} = \left\| Q^{dn}(s, \epsilon_{i^*}) - \mathrm{sg}\left[ Q^a(s, a_{i^*}) \right] \right\|_2^2.
\]

The regression-based learning and inference procedures are summarized in \Cref{alg:ours_learning,alg:ours_inference}. At test time, we select $i^* = \arg\max_{i \in \{1,\dots,N\}} Q^{dn}(s, \epsilon_i)$ and denoise only the selected candidate $\epsilon_{i^*}$ to obtain the final action $a_{i^*} = \pi_\theta(s, \epsilon_{i^*})$.

We evaluate \ours{} with two sampling-based methods in our experiments.

\textbf{\oursexpo{}. }
We implement \ours{} on top of EXPO, a high-performing online RL method that leverages a diffusion base policy combined with a lightweight edit policy. In this case, the selected denoised action $a_{i^*}$ serves as the base-policy proposal, and the downstream edit step follows EXPO unchanged.

\textbf{\oursidql{}. } We implement \ours{} on top of IDQL, which performs implicit action selection from a diffusion policy without an edit policy. In this case, the denoised action $a_{i^*}$ is executed directly.

\subsection{Computational profile} \label{ssec:comp_profile}

The key computational advantage of \ours{} over best-of-$N$ sampling is that it replaces $N$ full denoising rollouts with $N$ cheap noise-level critic evaluations followed by a single denoising rollout. Let $T$ denote the number of denoising steps, $N$ the number of candidates, and $F_{\mathrm{actor}}$, $F_{\mathrm{critic}}$ the FLOPs for a single forward pass of the actor and critic, respectively. Standard best-of-$N$ sampling denoises all $N$ candidates for $T$ steps each and then scores the resulting actions:
\[
  C_{\mathrm{BoN}} \;=\; T \cdot N \cdot F_{\mathrm{actor}} \;+\; N \cdot F_{\mathrm{critic}}.
\]
\ours{} instead scores all $N$ noise candidates with the noise critic and denoises only the top-scoring one:
\[
  C_{\oursplain{}} \;=\; N \cdot F_{\mathrm{critic}} \;+\; T \cdot F_{\mathrm{actor}}.
\]
The savings come from reducing the actor cost from $\mathcal{O}(TN)$ to $\mathcal{O}(T)$, eliminating $(N{-}1)$ full denoising passes. When $T = N$ and $F_{\mathrm{actor}} = F_{\mathrm{critic}} = F$, the cost reduces from $T(T{+}1) \cdot F$ to $2T \cdot F$---an approximate $T/2\times$ speedup.

In practice, $T$ typically ranges from 5 to 20 denoising steps. For example, EXPO~\citep{dong2025expo} uses $T{=}10$ and $N{=}8$. While one-step diffusion methods~\citep{liu2022flowstraightfastlearning,frans2025stepdiffusionshortcutmodels} attempt to distill multi-step generations into a single-step process, state-of-the-art diffusion models in robotics, image generation, and video generation~\citep{intelligence2025pi_,esser2024scalingrectifiedflowtransformers,gao2025seedream30technicalreport,wan2025} continue to use multi-step sampling in this range, making the cost reduction from \ours{} broadly applicable.

In our primary experiments, the actor and critic share the same architecture and FLOP count ($F_{\mathrm{actor}} = F_{\mathrm{critic}}$), consistent with baselines such as~\citep{dong2025expo} and recent work on large-scale RL~\citep{intelligence2025pi06vlalearnsexperience}. For our VLA experiments, the asymmetry is even more favorable: the critic has 20M parameters versus the actor's 3.3B, so $F_{\mathrm{critic}} \ll F_{\mathrm{actor}}$ and the $N$ noise-level evaluations add negligible overhead.

\begin{figure}[t]
  \centering
  \includegraphics[width=\linewidth]{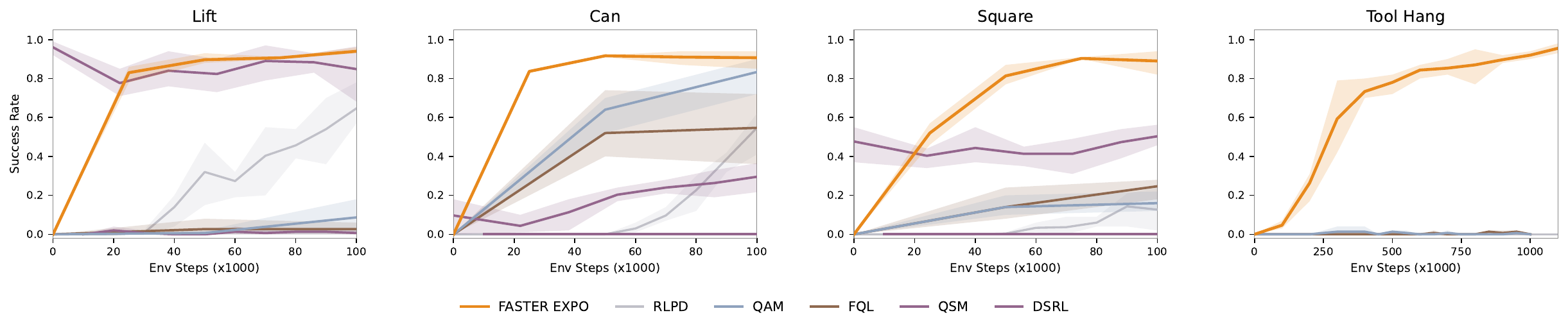}
  \includegraphics[width=\linewidth]{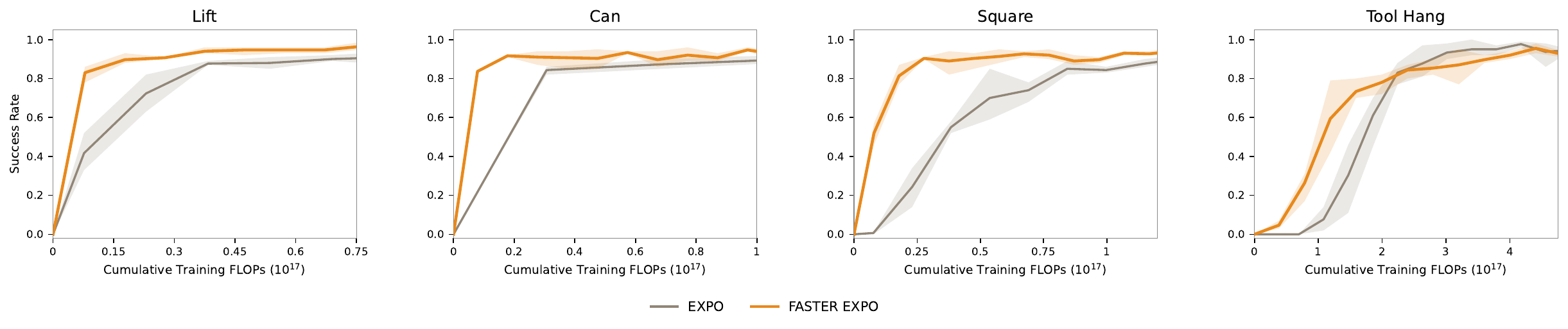}
  \caption{\footnotesize \textbf{Top: Success rates} of \ours{} and baselines in the online settings. \oursexpo{} outperforms strong baselines in sample efficiency. \textbf{Bottom: Compute comparisons} of \oursexpo{} and EXPO. \ours{} eliminates extra denoising during training and inference, yielding large FLOP reductions relative to the EXPO with comparable task
performance.}
  \label{fig:main_success}
\end{figure}

\begin{figure}[t]
  \centering
  \raisebox{0.3cm}{\includegraphics[width=0.22\linewidth]{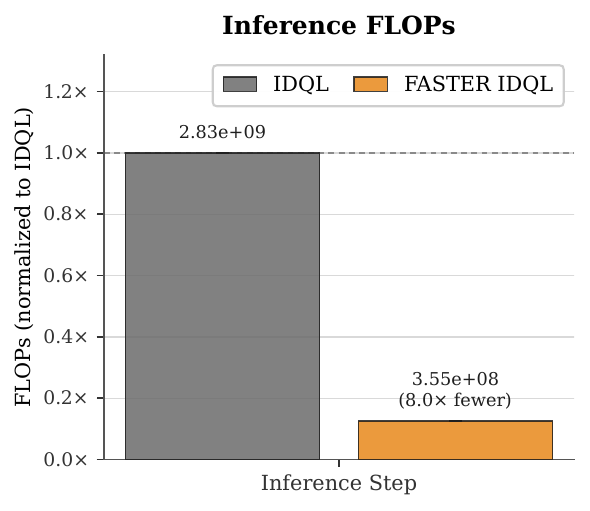}}
  \hspace{0.1cm}\includegraphics[width=0.75\linewidth]{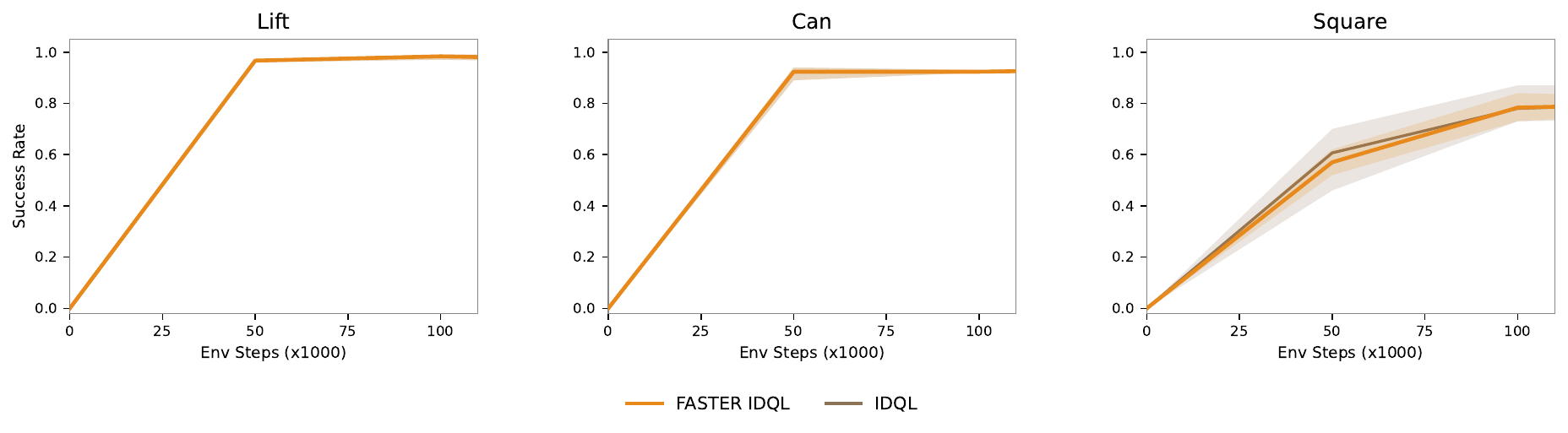}
  \caption{\footnotesize \textbf{Success rate and compute comparisons} of \oursidql{} and IDQL in the online setting. \ours{} can be applied to IDQL to eliminate extra denoising rollouts at inference while obtaining the same performance in success rates. (see \Cref{fig:main_success}).}
  \label{fig:main_flops}
\end{figure}

\begin{figure}[t]
  \centering
  \includegraphics[width=\linewidth]{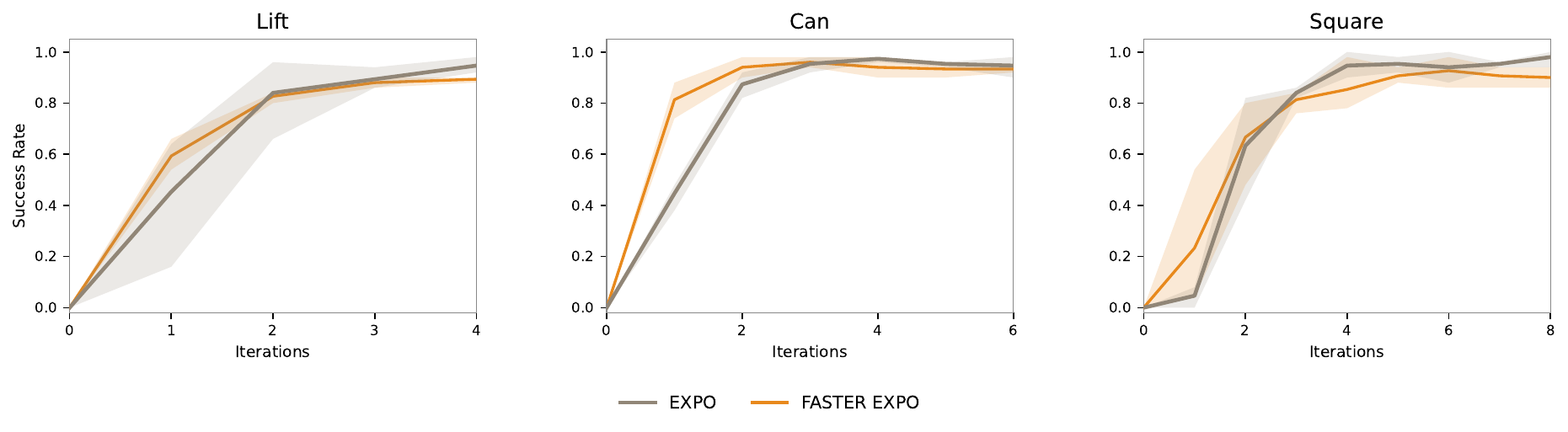}
  \includegraphics[width=\linewidth]{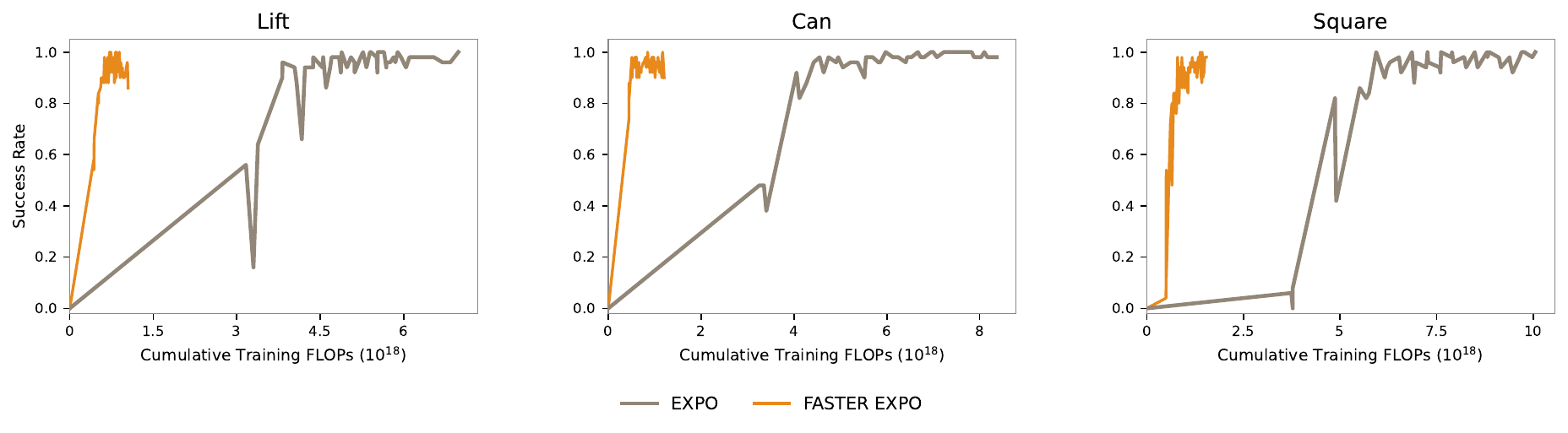}
  \caption{\footnotesize \textbf{Top: Success rate curves } of \oursexpo{} and EXPO in the batch-online setting. \ours{} matches the performance of EXPO in iterations. \textbf{Bottom: Compute comparisons } of \oursexpo{} and EXPO in the batch-online setting. Like in the online setting, in the batch-online setting \oursexpo{} yields a large FLOP reduction compared to EXPO from not needing to denoise all action samples. }
  \label{fig:discussion_batch_online_step_plots}
\end{figure}

%% file: experiments.tex
\section{Experiments} \label{sec:exp}

The goal of our experiments is to answer the following core questions:

\begin{enumerate}[start=1,label={(\bfseries Q\arabic*)}]
    \item How does \ours{} perform compared to state-of-the-art RL methods in both the online setting and the batch-online setting?
    \item Is \ours{} able to recover the performance gains of test-time scaling while reducing the computational cost?
    \item Can \ours{} be applied to a pretrained Vision-Language-Action (VLA) model? 
    \item What components of \ours{} are most important for performance? 
\end{enumerate}

\textbf{Environments. } We evaluate \ours{} on a set of 9 challenging manipulation tasks from Robomimic and LIBERO. All tasks feature a sparse reward indicating task completion. These environments involve controlling a 7-DoF robot arm to complete complex manipulation tasks. For Robomimic, we evaluate on \texttt{Lift}, \texttt{Can}, \texttt{Square}, and \texttt{Tool Hang}, which require picking up a block, picking up a can and placing it, inserting a tool onto a square peg, and hanging a tool on a rack, respectively. For LIBERO, we start from the pretrained \texttt{pi05\_libero} checkpoint from OpenPI~\citep{intelligence2025pi_}, which is trained on \texttt{libero\_goal}, \texttt{libero\_object}, \texttt{libero\_spatial}, and \texttt{libero\_10} but not \texttt{libero\_90}; we then select all held-out \texttt{libero\_90} tasks on which the original policy achieves 40--60\% success and run RL on those tasks without any offline data.

\textbf{Baselines. } We compare our method to prior state-of-the-art methods for online and batch-online \citep{dong2025mattersbatchonlinereinforcement} reinforcement learning.

\textbf{EXPO~\citep{dong2025expo}.} EXPO is an online RL method that jointly learns an expressive diffusion policy alongside a lightweight Gaussian edit policy that edits the actions sampled from the base policy toward a higher value distribution.

\textbf{IDQL~\citep{hansen2023idql}.} IDQL trains an expressive diffusion policy via imitation learning and uses implicit policy extraction by performing best-of-$N$ sampling, selecting the action that maximizes the $Q$-value.

\textbf{RLPD~\citep{ball2023efficient}.} RLPD is a highly sample-efficient algorithm that leverages prior data and samples from it for learning. RLPD uses a simpler Gaussian policy and has been shown to outperform many offline-to-online methods even without pretraining. For both evaluation settings, we run RLPD without offline pretraining.

\textbf{QSM~\citep{psenka2024qsm}.} QSM is an online RL method that trains diffusion policies by matching the diffusion loss to action gradients. QSM aims to avoid instability of value propagation to the expressive policy by incorporating losses to guide the denoising process.

\textbf{DSRL~\citep{wagenmaker2025latentrl}.} DSRL is an online RL method that adapts a frozen diffusion-based behavior cloning policy by performing RL over the initial noise seed used for sampling. DSRL learns a clipped Gaussian policy that predicts the initial noise.

\textbf{QAM~\citep{li2026qlearningadjointmatching}.} QAM is an online RL method that trains diffusion policies with adjoint matching. QAM aims to avoid instability of value propagation to the expressive policy by using action gradients to form a step-wise objective function that is free from unstable backpropagation.

\textbf{FQL~\citep{park2025flow}.} FQL uses a one-step flow policy to maximize the Q estimates learned with the standard TD objective. It also incorporates a behavioral regularization term toward a BC flow policy.

\subsection{How does \ours{} perform compared to state-of-the-art RL methods in both the online setting and the batch-online setting?}

We present the main results in \Cref{fig:main_success} and \Cref{fig:discussion_batch_online_step_plots}. Across both the online and batch-online settings, \oursexpo{} achieves the strongest overall performance among all baselines. In the online setting, \oursexpo{} significantly outperforms even highly sample-efficient methods such as RLPD. Methods that rely on action gradients to directly train an expressive policy, such as QSM and QAM, consistently struggle to match this performance. DSRL, which steers diffusion policies through the noise latent space, improves steadily but exhibits worse sample efficiency than methods that optimize directly in the action space. FQL, despite leveraging a one-step flow policy, consistently underperforms as behavioral regularization toward a BC policy can hinder exploration in the online setting.
In the batch-online setting, \oursexpo{} achieves the same performance as EXPO when comparing iterations, and outperforms EXPO in terms of compute efficiency. Overall, \ours{} achieves state-of-the-art performance across all evaluated methods.

\subsection{Is \ours{} able to recover the performance gains of test-time scaling while reducing the computational cost?} \label{ssec:main_performance}

We now evaluate whether the noise-level critic can identify high-quality noise seeds \emph{before} denoising, avoiding the redundant computation of fully denoising all $N$ candidates. We compare \oursexpo{} and \oursidql{} against their unfiltered counterparts, EXPO and IDQL, which rely on best-of-$N$ sampling at inference time. As shown in \Cref{fig:main_success} and \Cref{fig:main_flops}, both \ours{} variants achieve task success rates within the margin of error of their respective baselines across the evaluation suite in the online setting, confirming that the noise-level critic effectively recovers the performance gains of test-time scaling. Crucially, this comes at a fraction of the computational cost: as detailed in \Cref{ssec:comp_profile} and shown empirically in \Cref{fig:main_success}, \Cref{fig:main_flops} and \Cref{fig:ablation_vla_speed}, \ours{} reduces inference-time actor forward passes from $\mathcal{O}(TN)$ to $\mathcal{O}(T)$ by replacing $N$ full denoising rollouts with $N$ lightweight critic evaluations followed by a single denoising pass. Concretely, the update step goes from \SI{11.6}{\second} to \SI{2.5}{\second} and the inference step goes from \SI{566}{\milli\second} to \SI{335}{\milli\second}. These results demonstrate that the sample variance exploited by best-of-$N$ selection is largely determined at the noise level, and that \ours{} can capture this signal without incurring the cost of exhaustive denoising. We further discuss the computational tradeoffs between our methods and baselines in~\Cref{ssec:nondiffusion_comparison}.

\subsection{Can \ours{} be applied to a pretrained Vision-Language-Action (VLA) model?}
\begin{wrapfigure}{r}{0.5\textwidth}
    \vspace{-0.5cm}
    \centering
    \includegraphics[width=0.5\columnwidth]{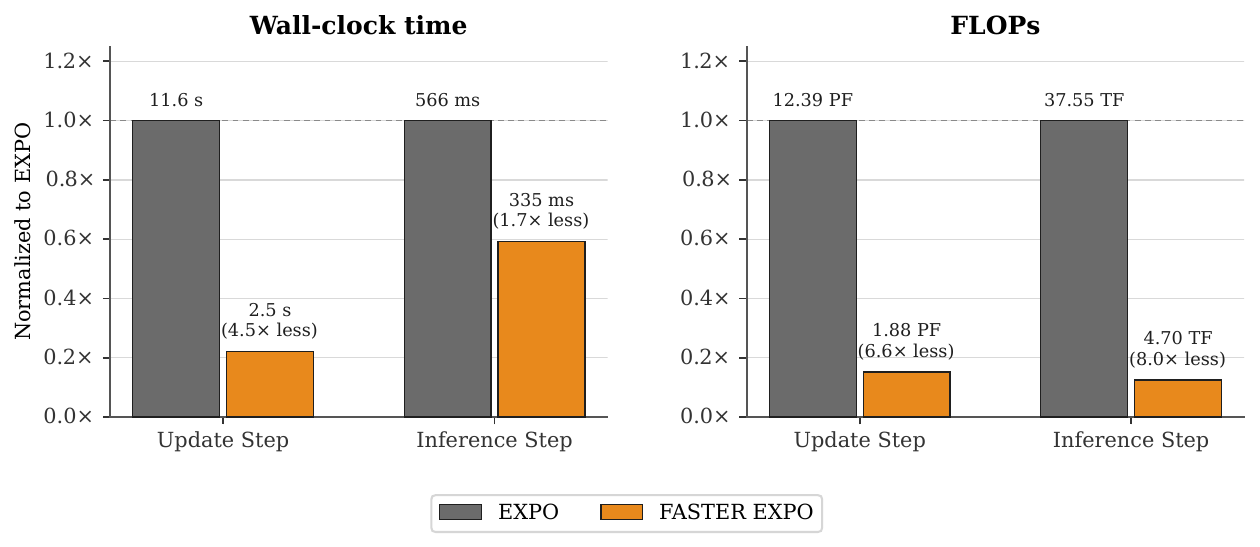}
    \vspace{-0.52cm}
    \caption{
        \footnotesize 
        \textbf{Training and inference timing} of \oursexpo{} compared to EXPO. \oursexpo{} achieves 1.7x improvement in inference time and 4.5x improvement in the update step time.
    }
    \vspace{-0.3cm}
    \label{fig:ablation_vla_speed}
\end{wrapfigure}

The computational burden of sampling-based methods is particularly acute as model scale increases; recent VLAs commonly reach 3B parameters, making the cost of denoising $N$ candidates at every environment step prohibitive for both training and deployment. We therefore evaluate whether \ours{} can be applied to a pretrained VLA to recover the benefits of best-of-$N$ sampling at substantially lower cost. Concretely, we apply \oursexpo{} to the 3.3B-parameter \texttt{pi05\_libero} checkpoint from OpenPI~\citep{intelligence2025pi_} and fine-tune it with online RL on 5 held-out \texttt{libero\_90} tasks, comparing against EXPO under identical conditions.

\textbf{Compute savings.} As shown in \Cref{fig:ablation_vla_speed}, \oursexpo{} reduces the per-step update time by approximately $4.5\times$ relative to EXPO, from roughly \SI{11.6}{\second} to \SI{2.5}{\second}. This speedup stems from the fact that \ours{} only denoises a single candidate through the 3.3B-parameter actor during each training step, whereas EXPO must denoise all $N$ candidates to score them with the env-level critic. At inference time, \oursexpo{} reduces latency from \SI{566}{\milli\second} to \SI{335}{\milli\second} per action, a $1.7\times$ speedup. The gains are even more pronounced in FLOPs: \oursexpo{} requires $4.70 \times 10^{12}$ inference FLOPs per step compared to $3.75 \times 10^{13}$ for EXPO---an $8\times$ reduction (\Cref{fig:ablation_vla_speed}). The theoretical tradeoff described in~\Cref{ssec:comp_profile} is especially favorable in the VLA setting because the noise-level critic (20M parameters) is far smaller than the actor (3.3B parameters), so scoring $N$ noise candidates adds negligible overhead relative to a single denoising pass.

\begin{figure}[t]
  \centering
  \includegraphics[width=\linewidth]{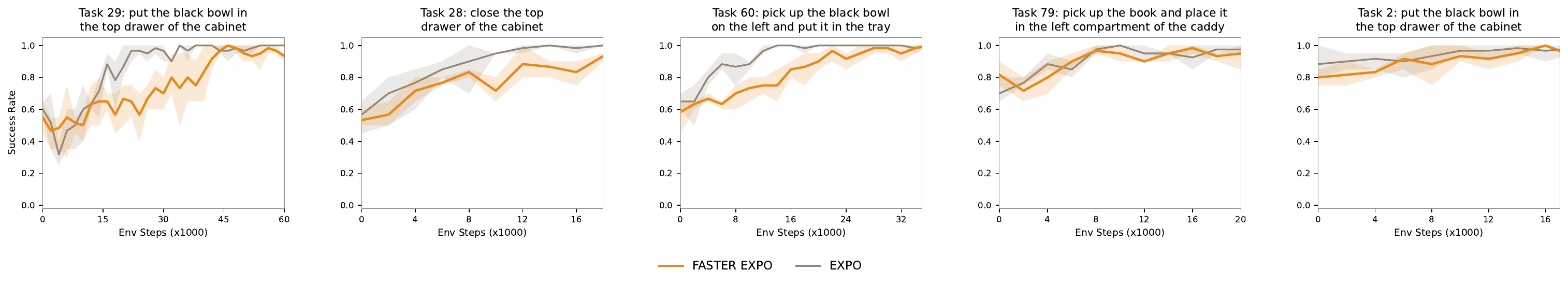}
  \includegraphics[width=\linewidth]{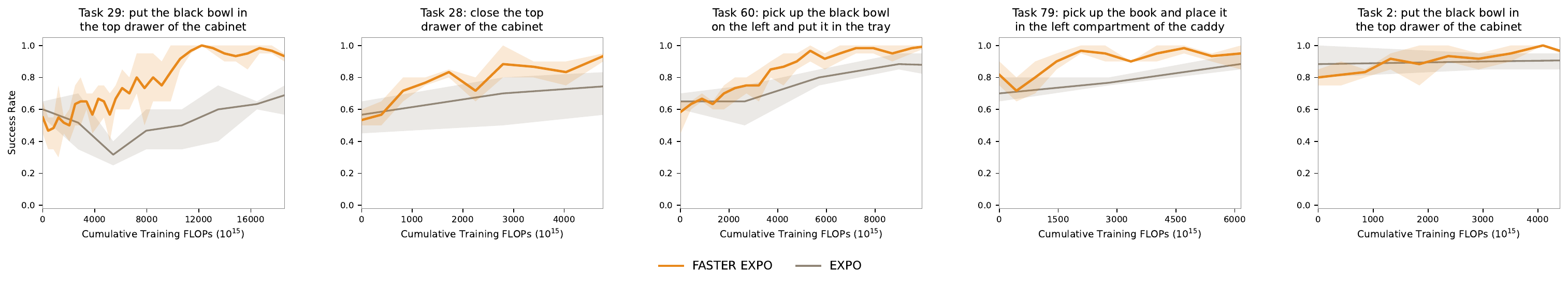}
  \caption{\footnotesize \textbf{\oursexpo{} compared to EXPO on top of $\pi_{0.5}$. } \textbf{Top: Performance with environment steps. } \oursexpo{} is competitive in performance compared to EXPO. \textbf{Bottom: Performance with FLOPs.} \oursexpo{} performs significantly better than EXPO under the same compute as \oursexpo{} chooses the best action sample without denoising all sampled actions in inference and training.   }
  \label{fig:vla_main}
\end{figure}

\textbf{Task performance.} Despite these substantial compute reductions, \oursexpo{} matches EXPO on the majority of the 5 evaluated tasks (\Cref{fig:vla_main}, top). Across the task suite, \oursexpo{} achieves comparable aggregate success rates to EXPO while requiring a fraction of the compute budget. These results demonstrate that \ours{} scales naturally to large pretrained VLAs, enabling efficient online RL fine-tuning without sacrificing the performance benefits of sampling-based action selection.

\subsection{What components of \ours{} are most important for performance?} To better understand how different pieces of \ours{} contribute to performance, we ablate over three key components: (1) the size of the denoise critic, (2) which step to filter with the denoise critic, and (3) learning the full filtering policy according to the MDP in Section~\ref{subsec:filter}. We present additional experiments on comparing \ours{} against distilling the value maximizing distribution in Section~\ref{app:exp} as another approach for reducing computational cost.

\begin{figure}[htbp]
    \centering
    \begin{minipage}[t]{0.48\textwidth}
        \centering
        \includegraphics[width=\textwidth]{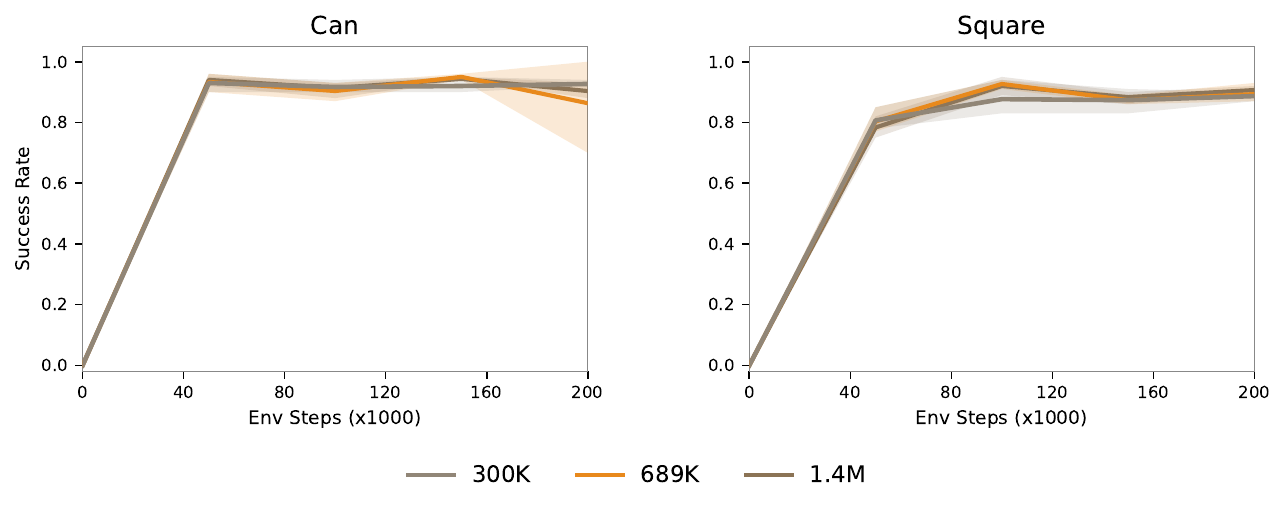}
        \caption{\footnotesize \textbf{Critic-size ablation} for \oursexpo{} on \texttt{can} and \texttt{square}. We compare filtering critics $Q^{dn}$ with parameter counts set to approximately $1.0\times$, $0.5\times$, and $0.25\times$ that of $Q^a$. We find that $Q^{dn}$ can be substantially smaller than $Q^a$ without degrading performance.}
        \label{fig:qf_scale_ablation}
    \end{minipage}
    \hfill
    \begin{minipage}[t]{0.48\textwidth}
        \centering
        \includegraphics[width=\textwidth]{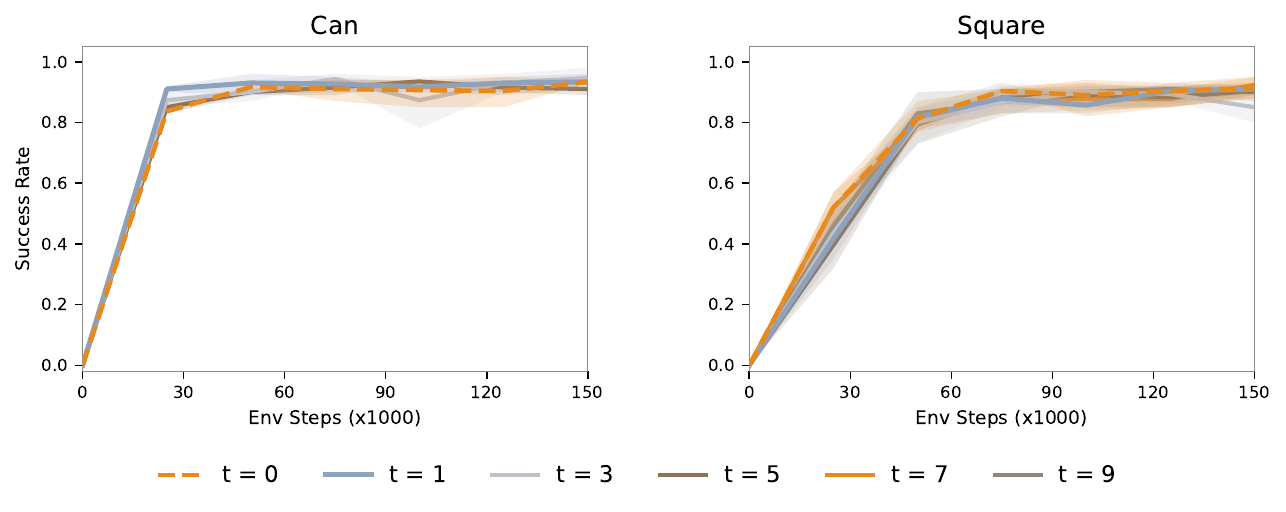}
        \caption{\footnotesize \textbf{Filtering step ablation. } \oursexpo{} achieves similar performance across different filtering steps, suggesting \ours{} does not require precise selection of the filtering step and filtering at the initial seed is effective.}
        \label{fig:discussion_ablation_filter_step}
    \end{minipage}

    \vspace{-0.3cm}

\end{figure}

\begin{wrapfigure}{r}{0.5\textwidth}
    
    \vspace{-0.3cm}
    \centering
    \includegraphics[width=0.5\columnwidth]{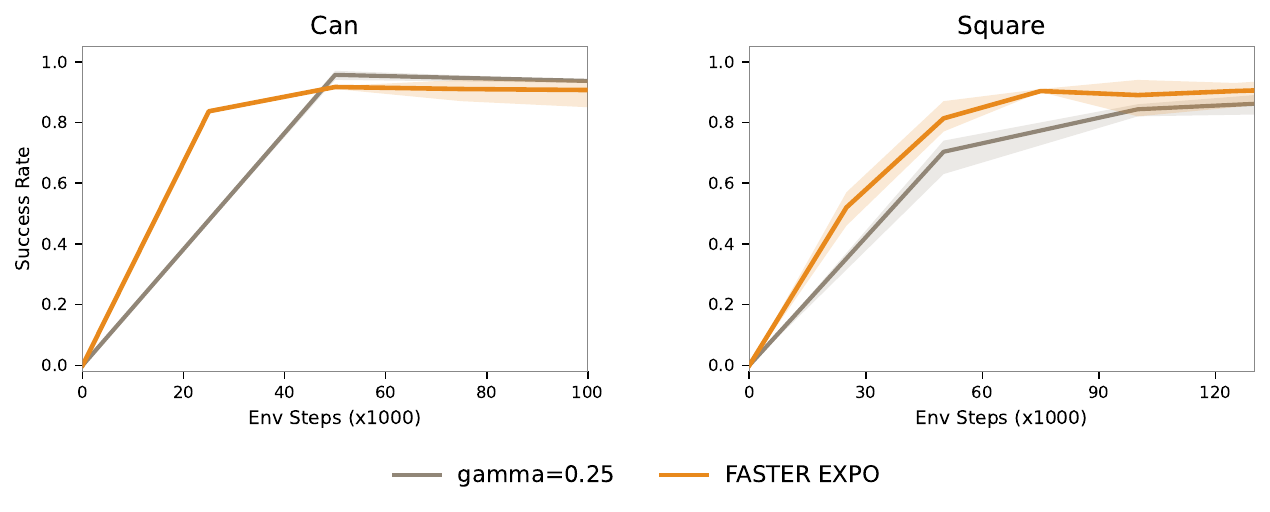}
    \caption{
        \footnotesize \textbf{Learned-filter ablation results. } Filtering at the initial seed performs comparably to learning the full filtering policy in the MDP. 
    }
    \vspace{-0.2cm}
    \label{fig:discussion_ablation_learned_filter}
\end{wrapfigure}

\textbf{How does the size of the denoise critic affect performance? } We investigate whether the capacity of the denoise critic is a significant factor in performance. Specifically, we compare architectures ranging from a smaller network (128 hidden units × 3 layers, 300K params) to a medium network (176 hidden units
× 3 layers, 689K params) to a larger network (256 hidden units × 3 layers, 1.4M params) on the Robomimic Can and Square tasks. As shown in Figure~\ref{fig:qf_scale_ablation}, performance remains largely consistent across all configurations, suggesting that \ours{} is robust to moderate variations in critic capacity and does not require careful tuning of this hyperparameter. Importantly, this suggests \ours{} can be used with a smaller denoise critic compared to action critic, which can greatly reduce computational cost in the case of large critic networks.

\textbf{How does the choice of filtering step affect performance? } In our experiments, we implement \ours{} by filtering only at the initial noise. A natural question is whether the specific denoising timestep at which the critic filter is applied has a meaningful impact on performance. We evaluate several choices of filtering step on the Robomimic Can and Square tasks. As reported in Figure~\ref{fig:discussion_ablation_filter_step}, performance is consistent across the range of steps considered, indicating that \ours{} does not require precise selection of the filtering timestep and is robust to this design choice; filtering at the initial seed achieves the best performance relative to the computation required.

\textbf{Policy learning with the full filtering MDP. } The complete formulation of \ours{} learns a filtering policy that can filter at any step of the denoising process. In our experiments, we reduce this multi-step filtering process to a single decision and implement filtering at the initial noise level, as this approach incurs minimal computational overhead while achieving performance on par with fully denoising all action candidates. Here, we present results for the general setting in which the filtering policy is learned to adaptively select the denoising step at which filtering occurs. In this case, $\gamma$ implicitly encourages the policy to terminate at earlier timesteps to save compute. Evaluating on the Robomimic Can and Square tasks, we find that the learned adaptive policy performs comparably to fixed filtering at the initial noise level. This suggests the single-step simplification of filtering at the initial seed serves as a good proxy for solving the full MDP.

%% file: discussion.tex
\section{Discussion} \label{sec:discussion}

In this work, we propose \ours{}, a method for obtaining the benefits of sampling-based test-time scaling without incurring its computational cost. By framing the denoising of multiple action candidates as an MDP, \ours{} learns a filtering policy that selects actions according to their downstream value. Despite these promising results, \ours{} has limitations. First, while \ours{} substantially improves upon its base algorithms in computational efficiency, it is not designed to improve sample efficiency. In our experiments, \oursexpo{} achieves better sample efficiency relative to strong baselines, owing largely to the inherent sample efficiency of EXPO, but does not surpass EXPO itself in this regard. Improving the sample efficiency of filtering-based approaches is an interesting direction for future work. Furthermore, \ours{} is designed for policy classes that use initial noise seeds, which encompass some of the most performant policy representations in contemporary reinforcement learning. Extending this to policy classes that lack such a seed structure is left to future work.

\nonsubmissiononly{
\section{Acknowledgments}
This work was supported by an NSF CAREER award and ONR grant N00014-22-1-2621. This work used the Delta system at the National Center for Supercomputing Applications [award OAC 2005572] through allocation CIS260152 from the Advanced Cyberinfrastructure Coordination Ecosystem: Services \& Support (ACCESS) program, which is supported by U.S. National Science Foundation grants \#2138259, \#2138286, \#2138307, \#2137603, and \#2138296.
}

%% file: appendix.tex
\section{Additional Experiments}
\label{app:exp}
\begin{figure}[H]
  \centering
  \includegraphics[width=\linewidth]{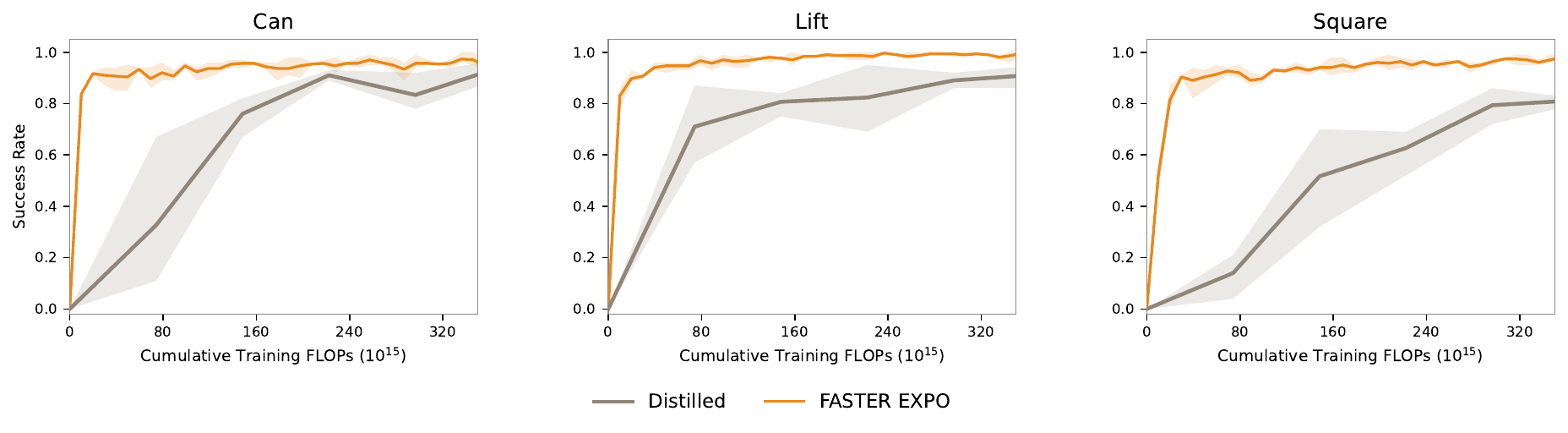}
  \caption{\footnotesize \textbf{Distillation ablation results. } Distilling the value maximizing distribution into a policy performs significantly worse compared to \ours{}. }
  \label{fig:discussion_ablation_distilled}
\end{figure}

A natural question is whether the best-of-$N$ sampling benefits observed in EXPO and IDQL can be recovered by a single policy trained via distillation. Concretely, an alternative to \ours{} would proceed as follows: sample $N$ noise candidates $\{\epsilon_i\}_{i=1}^N$ with $\epsilon_i \sim \mathcal{N}(0, I)$, denoise each to obtain actions $a_i = \pi_\theta(s, \epsilon_i)$, select the highest-scoring index $i^* = \arg\max_{i}\, Q^a(s, a_i)$, and distill the selected action $a_{i^*}$ into a new policy $\pi_\theta^\prime$ by supervising against $a_{i^*}$. We evaluate this distillation baseline in Figure~\ref{fig:discussion_ablation_distilled} and find that it performs substantially worse than \ours{}. We attribute this gap to the instability inherent in distillation: the distilled policy must continually chase a moving target defined by the $Q$-function, yielding a non-stationary training distribution that destabilizes learning. By contrast, \ours{} operates directly in noise space, framing selection as an easier filtering problem that is more amenable to stable optimization.

\section{Experiment Details}
\label{app:hyper}

\begin{table}[H]
    \centering
    \small
    \begin{tabular}{lcc}
    \toprule
        Hyperparameter & \oursexpo{} & \oursidql{} \\
    \midrule
        Actor / critic learning rate & $3\times 10^{-4}$ / $3\times 10^{-4}$ & $3\times 10^{-4}$ / $3\times 10^{-4}$ \\
        Batch size & $256$ & $256$ \\
        Discount & $0.99$ & $0.99$ \\
        Target update $\tau$ & $0.005$ & $0.005$ \\
        UTD ratio & $20$ & $20$ \\
        Candidates $N$ & $8$ & $8$ \\
        Candidates kept after filtering & $1$ & $1$ \\
        Denoising steps $T$ & $10$ & $100$ \\
        Filter temperature mode & z-score & z-score \\
        Expectile & --- & $0.8$ \\
        Edit scale & $0.15$ & --- \\
        \bottomrule
    \end{tabular}
    \vspace{0.2cm}
    \caption{\textbf{Hyperparameters} for the \oursexpo{} and \oursidql{} variants on Robomimic.}
    \label{tab:exp_details}
\end{table}

\textbf{Datasets and evaluation. } For the Robomimic experiments, all tasks use sparse binary rewards and report success rate. In the online setting, we do not pretrain on offline data, but we retain the same task conventions used in prior work to define task difficulty: \texttt{Lift} uses a 10-episode subset of the original dataset to make it more challenging, \texttt{Can} uses the multi-human dataset, and \texttt{Square} and \texttt{Tool Hang} use the standard proficient-human split. In the batch-online setting, we initialize training from these same demonstration sources and then continue interaction online. For the batch-online setting, we use $T=10$ denoising steps for \texttt{Lift} and \texttt{Can}, and $T=100$ denoising steps for \texttt{Square}. For the final rerun \oursexpo{} experiments, evaluation is performed every 50k environment steps on \texttt{Lift}, \texttt{Can}, and \texttt{Square}, and every 100k steps on \texttt{Tool Hang}; \oursidql{} uses the same evaluation protocol. Evaluation-time filtering with $Q^{dn}$ is greedy (temperature $0$) for all Robomimic tasks, while at training time we sample candidates from a softmax over z-score--normalized $Q$-values with temperature $1.0$ for \texttt{Lift}, \texttt{Can}, and \texttt{Square}, and $0.1$ for \texttt{Tool Hang}.

\textbf{VLA experiments. } For the pretrained VLA experiments, we follow the held-out LIBERO protocol described in the main text and fine-tune entirely online, without mixing in offline data. We use a batch size of $32$, UTD ratio $10$, and $100$k online training steps. We use $N=8$ candidates, retain a single candidate after filtering for \oursexpo{}, and denoise for $T=10$ steps. The VLA predicts action chunks over $10$ steps and we execute the entire chunk before replanning. Online optimization is performed at the episode level rather than every environment step: after each completed episode, we run three updates on newly sampled replay minibatches, with each update containing UTD 10 critic updates. To best utilize the replay data, we train on all sliding window chunks. For this setting, the critic is lightweight relative to the actor, so the additional cost of evaluating multiple noise candidates remains small compared to a full multi-candidate denoising pass. Both training-time and evaluation-time candidate selection use greedy sampling (temperature $0$), and we use an edit scale of $0.2$.

\textbf{Implementation Details. } We run experiments on a mix of NVIDIA A40, L40S, and H100 GPUs. All timing experiments are conducted on L40S GPUs. For update step timing, we consider the time taken for a single critic update, with other parameters as described in the prior paragraph. For inference step timing, we report BS=1 inference.

\section{Computational Profile}
\subsection{Comparison to non-diffusion-based methods}
\label{ssec:nondiffusion_comparison}
Our work focuses on capturing the benefits of sampling-based methods for diffusion models without incurring the  costs of these approaches. Diffusion and flow-based models are the dominant policy class in robotics, particularly in the real-world with VLAs~\cite{black2026pi0visionlanguageactionflowmodel,nvidia2025gr00tn1openfoundation} and WAMs~\cite{ye2026worldactionmodelszeroshot,pai2025mimicvideovideoactionmodelsgeneralizable}, hence our focus. Other methods, such as the Gaussian policies used in RLPD~\cite{ball2023efficient}, are generally more efficient due to lower typical parameter counts and no iterative computation—i.e., only a single forward step at inference. However, their limited expressivity makes them ill-suited for many real-world applications. Thus, this work focuses on the performance of diffusion methods with the same parameter count, which we believe to be the most apt comparison.